\documentclass[sigconf,nonacm]{acmart}
\setcopyright{none}
\AtBeginDocument{%
  }

\usepackage{pdfpages}
\usepackage{arydshln}
\usepackage{makecell}
\usepackage{graphicx}

\begin{document}

\title{EventGPT: Capturing Player Impact from Team Action Sequences Using GPT-Based Framework}

\author{Minho Lee}
\authornotemark[1]
\affiliation{%
  \institution{Institute for Sports and Preventive Medicine, Saarland University}
  \city{Saarbrücken}
  \country{Germany}
}
\email{minho.lee@uni-saarland.de}

\author{Miru Hong}
\authornote{Both authors contributed equally to this research.}
\affiliation{%
  \institution{Department of Artificial Intelligence, University of Seoul}
  \city{Seoul}
  \country{Republic of Korea}
}
\email{mirunoyume@uos.ac.kr}

\author{Geonhee Jo}
\affiliation{%
  \institution{Department of Artificial Intelligence, University of Seoul}
  \city{Seoul}
  \country{Republic of Korea}
}
\email{geonhee@uos.ac.kr}

\author{Jae-Hee So}
\affiliation{%
 \institution{Bank of Korea}
 \city{Seoul}
 \country{Republic of Korea}
}
\email{jayso@bok.or.kr}

\author{Pascal Bauer}
\affiliation{%
  \institution{Chair for Sports Analytics, Saarland University}
  \city{Saarbrücken}
  \country{Germany}
}
\affiliation{%
  \institution{Deutscher Fussball-Bund(DFB)}
  \city{Frankfurt}
  \country{Germany}
}
\email{pascal.bauer@uni-saarland.de}

\author{Sang-Ki Ko}
\authornote{Corresponding author}
\affiliation{%
  \institution{Department of Artificial Intelligence, University of Seoul}
  \city{Seoul}
  \country{Republic of Korea}
  }
\email{sangkiko@uos.ac.kr}

\renewcommand{\shortauthors}{Hong et al.}

\begin{abstract}
  Transfers play a pivotal role in shaping a football club’s success, yet forecasting whether a transfer will succeed remains difficult due to the strong context-dependence of on-field performance. Existing evaluation practices often rely on static summary statistics or post-hoc value models, which fail to capture how a player’s contribution adapts to a new tactical environment or different teammates. To address this gap, we introduce EventGPT, a player-conditioned, value-aware next-event prediction model built on a GPT-style autoregressive transformer. Our model treats match play as a sequence of discrete tokens, jointly learning to predict the next on-ball action’s type, location, timing, and its estimated residual On-Ball Value (rOBV) based on the preceding context and player identity. A key contribution of this framework is the ability to perform counterfactual simulations. By substituting learned player embeddings into new event sequences, we can simulate how a player’s behavioral distribution and value profile would change when placed in a different team or tactical structure. Evaluated on five seasons of Premier League event data, EventGPT outperforms existing sequence-based baselines in next-event prediction accuracy and spatial precision. Furthermore, we demonstrate the model's practical utility for transfer analysis through case studies—such as comparing striker performance across different systems and identifying stylistic replacements for specific roles—showing that our approach provides a principled method for evaluating transfer fit.
\end{abstract}



\maketitle

\section{Introduction}
Transfers play a pivotal role in shaping a football club’s sporting and financial future, often determining its long-term success or failure. The importance of the market keeps rising: according to FIFA, the 2025 mid-year window alone saw a record USD 9.76 billion in transfer fees, with England the top-spending association ~\cite{FIFA2025Transfers}. Nevertheless, forecasting whether a transfer will succeed remains difficult. Post-transfer performance is influenced by multiple interacting factors, including the competitive level of the destination league, the tactical philosophy of the new coaching staff, the player’s assigned role within team structure, squad hierarchy and chemistry, physical adaptation and workload patterns, and even cultural and environmental acclimatization~\cite{Berlinschi2013, Jarjabka2024}. These elements interact in ways that are nonlinear and often opaque, making simple evaluation based on past performance insufficient.

Among these challenges, a particularly critical issue is the strong context-dependence of on-field contribution. The same player can produce markedly different outcomes depending on the surrounding tactical structure, the behaviors of teammates, the intensity and structure of the opponent’s pressing, and situational match states (such as leading vs. trailing or transitions vs. settled play). Yet, current scouting and evaluation practices still rely heavily on static summary statistics or highlight-based assessments, which capture ability but not how that ability translates under new conditions. This gap highlights the need for models that explicitly learn and evaluate performance as a function of playing context, rather than treating actions as context-free indicators of individual quality.

A substantial line of research has focused on evaluating how much an on-ball action contributes to a team’s chance of scoring or preventing goals. Decroos et al.~\cite{DecroosVAEP2021} introduced VAEP, a framework that estimates the change in scoring and conceding probabilities induced by each action, demonstrating that value is not inherent to the action itself but emerges from its context within the game state. Similarly, Expected Threat (xT) introduced a spatial scoring-potential surface, emphasizing how ball progression into certain pitch zones increases goal probability ~\cite{Singh2019xT}. On-Ball Value (OBV) assigns value to each on-ball action based on how it changes the expected outcome of the current possession sequence ~\cite{StatsBomb2021OBV}. More recently, Expected Possession Value (EPV) extends these ideas by estimating the value of entire possessions and modeling how present states imply future scoring chances~\cite{Fernandez2019EPV}. These methods have been influential in highlighting that action impact is fundamentally contextual. However, value estimation is typically applied as a post-hoc layer on completed event sequences; the value of an action is evaluated after it happens, rather than co-learned with the sequential process that generates actions. As a result, these models capture what was valuable, but not how that value would change when a player operates in a different tactical environment.

Another line of research has focused on learning player representations that capture style, role, and behavioral tendencies. Early approaches constructed player profiles from aggregated event statistics or possession chain summaries, while later work introduced embedding-based methods that learn latent representations from match event sequences or network structures~\cite{Pappalardo2019, DecroosVector2019}. Such embeddings support tasks such as similarity search, role classification, and recruitment shortlist generation, and they have proven effective for identifying players with comparable tactical functions. More recent work incorporates relational structure or interaction patterns, for example, by modeling passing networks or co-occurring on-ball behaviors, in order to represent how a player’s contributions depend on teammates and system shape ~\cite{YilmazEmbedding2022, Rovshitz2024TransformerSoccer, adjileye2024risingballer}. However, these representations are generally static: they summarize how a player has behaved in their previous context, but they do not model how that behavior would adapt when the surrounding tactical environment changes. Consequently, while embedding approaches help answer the question of who a player is, they do not address the core transfer-fit question of what this player would do in a different team, under different roles and constraints.

Recent work has approached match play from a sequential modeling perspective, treating event streams as structured sequences rather than isolated observations. Early next-event prediction models, such as Seq2Event~\cite{Simpson2022Seq2event}, demonstrated that the unfolding flow of possession can be captured autoregressively, learning to predict the next action based on preceding context and thereby revealing regularities in tactical organization. Extending this idea, neural marked spatio-temporal point process models (NMSTPP) ~\cite{Yeung2025NMSTPP} jointly model the type, location, and timing of future actions, emphasizing that match dynamics are fundamentally spatio-temporal rather than purely categorical. More recently, Large Event Models (LEM) ~\cite{MendesNeves2026LEM} have advanced this autoregressive view toward a generative framing in which full match progressions can be simulated from arbitrary game states, moving toward foundation-model-style applications such as scenario exploration and tactical evaluation. However, existing sequence-based and large-event models typically do not explicitly condition on player identity and generally do not jointly learn action value within the generative process. As a result, they describe how play unfolds in aggregate, but do not account for how the presence of a specific player would alter the trajectory and value of actions within a new tactical environment—precisely the core problem in evaluating transfer fit.

\begin{figure*}[t]
    \centering
    \includegraphics[width=0.8\linewidth]{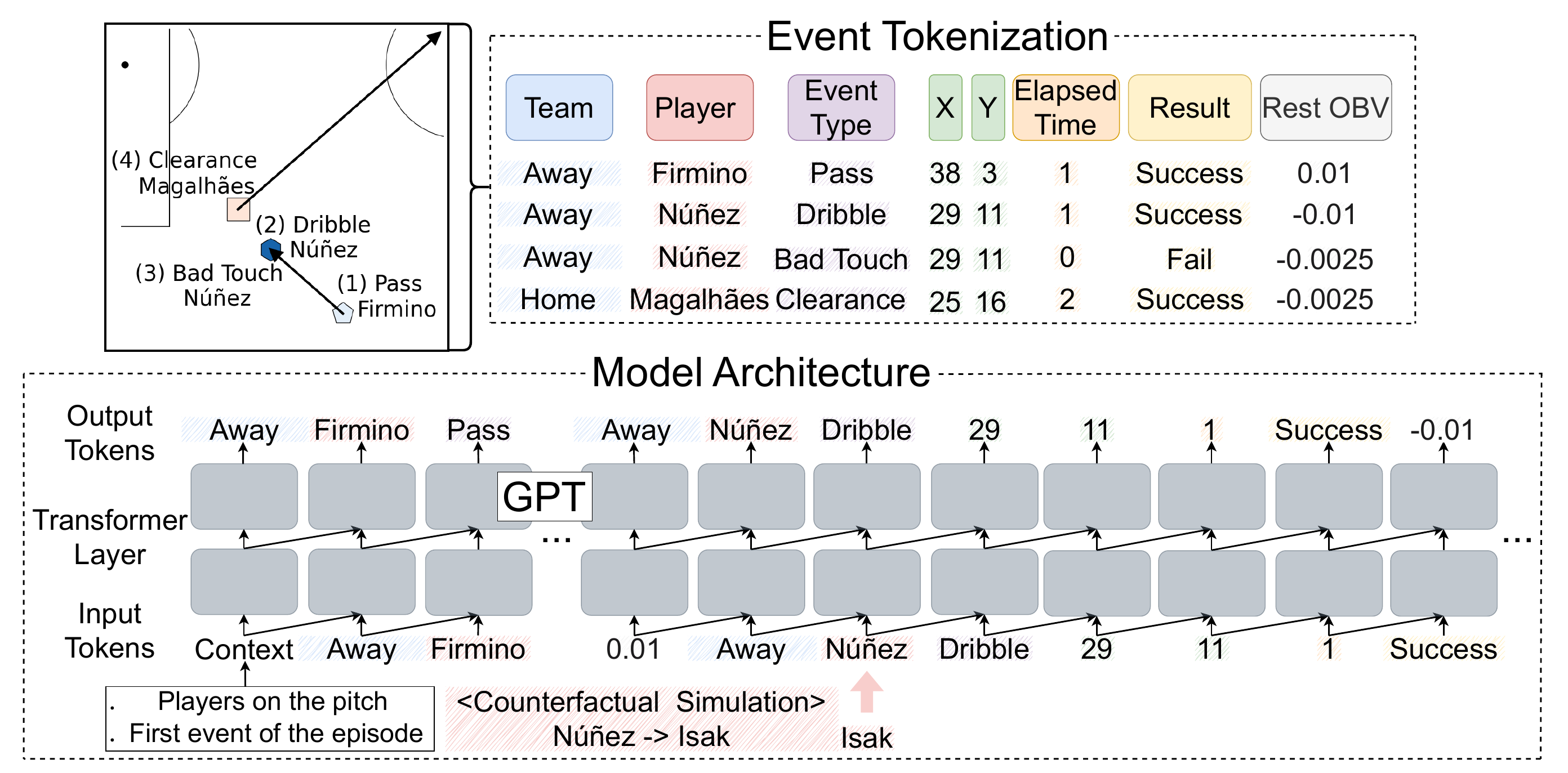}
    \caption{Overview of the EventGPT framework. Our nanoGPT-based Transformer model autoregressively predicts event tokens, enabling counterfactual `what-if' simulations. For instance, replacing Darwin N\'u\~nez with Alexander Isak could alter actions (e.g., pass/shot) or modify the same action with a different location, outcome, or on-ball value (OBV). A video example of this sequence can be seen at the 23-second mark in the \href{https://www.youtube.com/watch?v=y9wH5s0C33w}{\textit{linked video clip}}.}
    \label{fig:overview}
\end{figure*}

To address this gap, we introduce EventGPT, a player-conditioned, value-aware next-event prediction model built on a GPT-style autoregressive transformer. Figure 1 provides an overview of the EventGPT framework, illustrating how our model autoregressively predicts event tokens, enabling counterfactual 'what-if' simulations. The model jointly predicts the next on-ball action and its estimated on-ball value, with player identity provided as an input through learned embeddings that influence the prediction of all downstream event attributes, including action type, spatial location, timing, outcome, and action-level value. Crucially, the learned player embeddings can be substituted into new team event sequences, enabling counterfactual simulation of how a player’s action distribution and value profile would change when placed in a different tactical environment. This provides a principled framework for context-aware transfer fit evaluation, which existing action-value models, player representations, and sequence-based event models do not directly support.

Taken together, our approach contributes the following:
\begin{itemize}
    \item A player-conditioned next-event prediction framework that jointly models on-ball action value (OBV), enabling evaluation of player impact in transfer scenarios.
    \item A GPT-style autoregressive architecture that leverages large-sequence modeling capabilities developed in language models, adapted directly to football event streams.
    \item A standardized event representation pipeline using SPADL formatting and episode-based segmentation, ensuring consistent and transferable sequential input structure.
\end{itemize}

We evaluate our model on five seasons of Premier League event data (2020/21–2024/25), using SPADL-formatted and episode-segmented match sequences. Across standard next-event prediction benchmarks, our method outperforms baseline models, including non–player-conditioned and non–value-aware architectures. Beyond predictive accuracy, we demonstrate the model’s practical utility in transfer analysis through counterfactual player substitution experiments, showing that the learned player-conditioned and value-aware representations can meaningfully approximate how a player’s on-ball contribution profile would change when moving to a different tactical environment.


\section{Methodology}
Our model is a decoder-only Transformer based on the GPT architecture, modeling football match
play as a sequence of discrete-event tokens. All event attributes—including player ID, action type,
spatial coordinates, timing, outcome, and action-level value (e.g., OBV)—are represented within a
unified vocabulary and processed through a single embedding layer. Match episodes are encoded
as fixed-length sequences composed of a context block listing the 22 on-pitch players and game
states, followed by autoregressively predicted event tokens. The model is trained with next-token
prediction using teacher forcing. Further details are described in the subsections below.
\subsection{Event Representation and Tokenization}

Each episode is represented as a fixed-length sequence of discrete tokens. We adopt the SPADL event representation to standardize action semantics. A unified vocabulary assigns non-overlapping token index ranges to event attributes, player identities, and special context markers. Continuous attributes such as spatial coordinates, elapsed time, and OBV are discretized to bounded integer ranges for compatibility with the shared embedding layer.

Matches are segmented into episodes, defined as contiguous sequences of play during which the set of 22 on-pitch players remains unchanged. A new episode begins whenever play is reset (e.g., set-pieces, goals, or period transitions) or when substitutions or dismissals alter the active players. This segmentation ensures that each episode corresponds to a consistent tactical and personnel context, which is necessary for player-conditioned prediction and masking.
Every episode begins with a context block that encodes the match state at the episode start, including:
\[
c = (pID_{1:22}, \mathit{minute}, h_{g^{\prime}}, a_{g^{\prime}}, h_{r^{\prime}}, a_{r^{\prime}}, h_{y^{\prime}}, a_{y^{\prime}}),
\]
where $pID_{1:22}$ denotes the on-pitch players' ID, $\mathit{minute}$ is the current match minute, $h_{g^{\prime}}, a_{g^{\prime}}$ are cumulative goals, and $h_r, a_r, h_y, a_y$ denote cumulative red/yellow cards for the home and away teams. This ensures that the model conditions event prediction on both who is on the pitch and the current match situation.

Following the context block, up to 100 events are appended, each encoded as a sequence of tokens describing their attributes. Episodes exceeding this length are truncated to the most recent events, and shorter sequences are right-padded for batching.

\subsection{Autoregressive Next-Event Prediction}

The model generates events token-by-token in an autoregressive manner. Each event is represented not by a single symbol but by a sequence of tokens corresponding to its component attributes:
\[
v_t=(h_t, e_t, x_t, y_t, \delta_t, o_t,{\mathrm rOBV}_t)
\]
where $h_t$ indicates the acting team (home or away), $e_t$ is the event type, and $(x_t,y_t)$ are the discretized spatial coordinates. The term $\delta_t$ represents the elapsed time since the previous event, and $o_t$ indicates whether the action was successful. The final token ${{\rm rOBV}_t}$ denotes the residual on-ball value contribution of the player ${p_t}$from time ${t}$ until the end of the episode:
\[
\mathrm{rOBV}_{t} = E \left[ \sum_{\tau=t}^{T_{\mathrm{episode}}} \mathrm{OBV}_{\tau} \mid \text{state at } t, \text{player } p_{t} \right],
\]

where ${p_t}$ is the player ID token, provided only as part of the input context. Unlike immediate action value estimates, ${\mathrm{rOBV}_{t}}$ reflects how a player’s involvement influences the future progression of the episode. This allows the model to learn how replacing a player affects not only the current time, but also the subsequent structure and outcome of the possession sequence, enabling realistic transfer-fit simulations.

Importantly, the player identity ${p_t}$ conditions the prediction but is never itself predicted; this enables controlled counterfactual substitution by simply replacing the identity token while holding the surrounding context fixed.
Given a token sequence $(v_1, v_2, ... v_t)$, the model predicts the distribution of the next token 
\[
P(v_{t+1} | c, p_{1:t'}, v_{1:t}).
\]
Training is performed using teacher forcing and minimizing the autoregressive next-token prediction loss:
\[
L = - \sum_{t=1}^{T} \log P(v_{t+1} \mid c, p_{1:t} v_{1:t})
\]
This training objective jointly supervises event sequencing, player-conditioned action selection, and future value estimation within a unified autoregressive process.

\subsection{Model Architecture}

 follows a decoder-only Transformer architecture directly adapted from the NanoGPT implementation of the GPT model by Karpathy ~\cite{Karpathy2022nanoGPT}. The network consists of a stack of causal self-attention decoder blocks, each comprising multi-head attention and position-wise feedforward layers with residual connections and layer normalization. Token embeddings are applied through a single shared embedding matrix over the unified vocabulary described in Section 2.1, and learned positional embeddings are added to preserve event order within each episode. The final layer ties the output projection weights to the input embedding matrix, enabling efficient next-token prediction across all event attributes.

Since football event streams share structural similarity with natural language sequences—where meaning is conveyed through ordered symbolic tokens—the NanoGPT-style autoregressive Transformer provides a natural architectural fit while remaining computationally efficient and scalable to large datasets. In contrast to multi-branch architectures that separately embed spatial, temporal, and categorical event features, our model represents all components as tokens, allowing the Transformer to learn their interactions through attention rather than manual feature fusion.

\subsection{Hypothetical Player Substitution for Transfer Simulation}

The learned player embeddings encode how individual players influence both the local structure of their actions and the future progression of an episode, as captured through the residual on-ball value target $\mathrm{rOBV}_{t}$. To evaluate how a player may perform in a different tactical environment, we perform counterfactual player substitution by modifying only the target player ID in the episode.
Given an episode with a context block $c$, we replace a player $p_i$ with another player $p_j$ while leaving all other tokens unchanged to form a modified context block $c^{\prime}$. The model is then used to re-evaluate the residual value predictions $\mathrm{rOBV}_{t}$ on the unchanged event sequence, conditioning the value estimates on the substituted player identity. This isolates how the player’s presence would alter the expected future contribution within the same tactical and match-state environment.

\section{Experiments}

\subsection{Dataset}

Our model was evaluated on five seasons of Premier League event data (2019/20–2023/24), standardized using the SPADL event representation. Matches are segmented into episodes such that each episode contains a consistent set of 22 on-pitch players, as described in Section 2.1. Episodes are further processed into fixed-length token sequences of up to 100 events, with truncation for longer sequences and right-padding for shorter ones. The model is trained on data from the 2020/21, 2021/22 and 2022/23 seasons, as well as the first half of the 2023/24 and 2024/25 seasons; the remaining data is held out for evaluation. Detailed dataset statistics can be found in Table~1.
\begin{table}[t]
\centering
\caption{Summary of the dataset used in experiments.}
\label{tab:dataset_stats}
\begin{tabular}{cccc}
\toprule
Property & Value \\
\midrule
Seasons & 2020/21 - 2024/25 \\
Total Number of Matches & 1{,}900 \\
Total Number of Episodes & 173{,}951 \\
Average Number of Events per Episode & 22.48 \\
Total Number of Players & 1{,}221 \\
\bottomrule
\end{tabular}
\end{table}

\subsection{Baselines}

We compare our model against sequence-based event prediction approaches derived from prior work on next-event modeling. However, existing models such as Seq2Event and NMSTPP are not directly comparable in their original form, because they predict only coarse-grained action types (e.g., pass, dribble, shot) and do not model additional event attributes such as player identity, spatial coordinates, temporal intervals, or action value. Furthermore, these models typically do not incorporate player conditioning, which is central to our evaluation of transfer fit. To enable a fair comparison, we adapt the Transformer backbone introduced in NMSTPP to our output formulation: preserving the architecture while replacing the output prediction heads with those that match our multi-attribute event representation. We denote this baseline as LEM Transformer.

\subsection{Evaluation Metrics}

We evaluate model performance across discrete and continuous event attributes using classification accuracy and mean absolute error (MAE), respectively.
\begin{itemize}
    \item \textbf{Home/Away Team Indicator} ($h_{t}$): Accuracy
    \item \textbf{Event Type} ($e_{t}$): Accuracy
    \item \textbf{Spatial Coordinates} ($x_{t}, y_{t}$): MAE (measured in meters after rescaling)
    \item \textbf{Elapsed Time} ($\delta_{t}$): MAE
    \item \textbf{Action Success Indicator} ($o_{t}$): Accuracy
    \item \textbf{Residual On-Ball Value} ($\mathrm{rOBV}_{t}$): MAE
\end{itemize}
Accuracy is used for categorical prediction tasks, while continuous outputs are evaluated using MAE computed on the original (non-discretized) scale.

\subsection{Main Results}

Table 2 reports next-event prediction performance across categorical (home/away indicator, player identity, event type) and continuous (location, elapsed time, rOBV) attributes. Compared to the original LEM and the reconfigured LEM Transformer baseline, our proposed model achieves substantially higher accuracy in event prediction, as well as lower error in spatial and value estimation.
\begin{table}[t]
\centering
\caption{Next-event prediction performance across baselines. Bold indicates top results, and underline indicates second-best results.}
\label{tab:model_comparison}

\resizebox{\linewidth}{!}{
\begin{tabular}{lccccccc}
\toprule
Model & h↑ & e↑ & x↓ & y↓ & t↓ & a↑ & rOBV↓ \\
\midrule
LEM              & 85.20\% & 74.07\% & 9.01 & 8.11 & \underline{1.37} & \underline{90.51}\% & 0.014 \\
LEM Transformer  & \textbf{96.05\%} & \underline{80.42}\% & \underline{7.15} & \underline{7.08} & 1.40 & 86.92\% & \textbf{0.008} \\
EventGPT         & \underline{94.12}\% & \textbf{82.91\%} & \textbf{4.30} & \textbf{4.31} & \textbf{1.11} & \textbf{92.87\%} & \underline{0.009} \\
\bottomrule
\end{tabular}
}
\end{table}

Notably, our model improves event type prediction accuracy compared to both baselines (82.91\% vs. 80.42\% for the LEM Transformer and 74.07\% for LEM), indicating that conditioning on player identity and episode context leads to more accurate modeling of action selection. In addition, our model substantially reduces spatial prediction error, demonstrating improved understanding of how play unfolds on the field. While the LEM Transformer obtains the lowest rOBV error (0.008), EventGPT achieves a competitive second-best result (0.009).

both event realism and contribution estimation, supporting their use for downstream transfer-fit simulation.

\section{Applications}

\subsection{Player Embedding for Similar Player Retrieval}

To examine whether the learned player embeddings capture meaningful clustering structures 

\begin{figure}
    \centering
    \includegraphics[width=1.\linewidth]{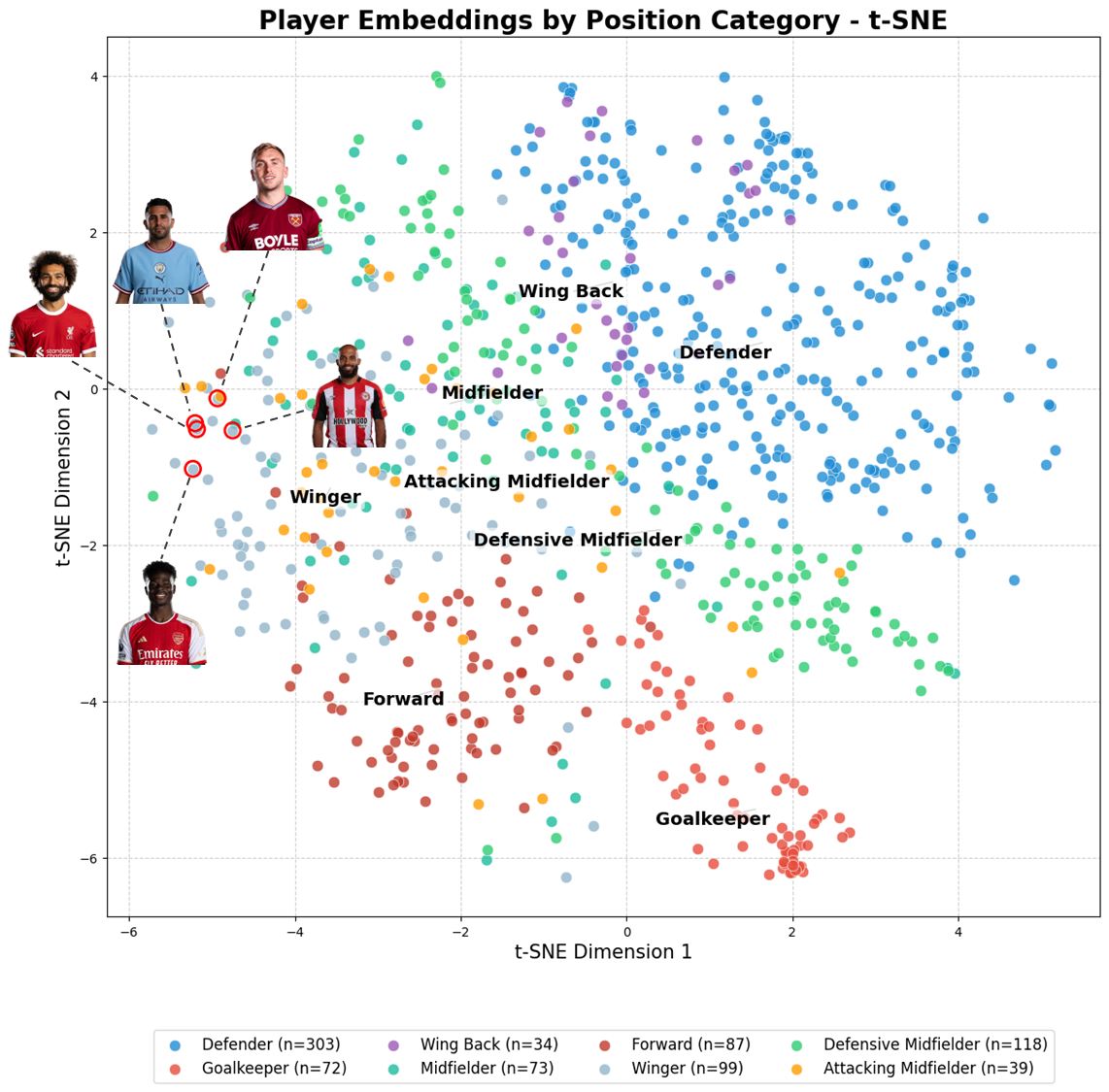}
    \caption{Player Embeddings by Position Category. This visualization demonstrates how functional tendencies are encoded through sequential event patterns and learned player-conditioned residual on-ball value (rOBVt) dynamics, even without using positional labels during training.}
    \label{fig:placeholder}
\end{figure}

related to on-field roles, we project the player embedding vectors into two dimensions using t-Distributed Stochastic Neighbor Embedding (t-SNE), as illustrated in Figure 2. t-SNE is a non-linear visualization technique adept at preserving high-dimensional local neighborhood structures, making it highly effective for revealing distinct player archetypes. Each point corresponds to a single player, and colors denote coarse positional categories (e.g., defenders, wing backs, central midfielders, attacking midfielders, wingers, and goalkeepers).

In Figure 2, we observe clear, role-consistent clusters. Defensive players, particularly center-backs and full-backs, form a dense and distinct region, reflecting a high similarity in their sequential event patterns. Wing-backs emerge in an intermediate space between defenders and wingers, consistent with their mixed responsibilities in buildup and wide progression. Defensive midfielders and midfielders appear adjacent, acting as a bridge between the defensive cluster and the more advanced attacking players. Meanwhile, attacking midfielders, wingers, and forwards form several distinct sub-clusters in a separate region, highlighting the model's ability to differentiate functional roles within the final third.

Notably, this structure arises without using any positional labels during training; the model infers these similarities purely from sequential event patterns and the learned player-conditioned rOBV value dynamics. This suggests that the embedding space encodes functional tendencies rather than nominal position, supporting its use in transfer-fit evaluation and similarity-based recruitment.

\subsection{Hypothetical Transfer Simulation}

A key component for validating our model is counterfactual simulation. For example, we can simulate Rasmus Højlund within the exact tactical context of Erling Haaland at Manchester City to project his behavioral statistics, or conversely, place Haaland in Højlund’s environment. Such counterfactual scenarios allow us to evaluate whether the model produces consistent and realistic outcomes conditioned on the interaction between a player's learned embedding and a given tactical context.

The simulation procedure begins by collecting all event sequences involving the target player during the season. For each episode, we keep the observed event sequence fixed and condition the model on the corresponding tactical context (on-pitch players and match state). We then substitute the acting player identity with that of another player and re-evaluate the residual value predictions under the modified context.

As described in Section 2, the model autoregressively generates the next event token sequence. In the transfer simulation setting, the home/away indicator is already included in the context block and is thus used only as conditioning information rather than a prediction target. The generation is repeated multiple times for each episode to obtain a distribution of plausible outcomes.

To compute the mean expected value from these ${N}$ samples, we use a position-specific aggregation strategy. For most roles (e.g., midfielders and defenders), which are characterized by relatively low-variance action profiles, the standard arithmetic mean of all ${N}$ samples serves as a robust estimator. A key exception is made for attackers. This position exhibits a fundamentally different, high-variance OBV distribution, with a low frequency of high-value outcomes (e.g., goals). In this case, a simple average would be disproportionately influenced by the frequent low-value outcomes and would therefore underrepresent the player’s potential impact. To account for this, we employ a truncated mean—specifically, the average of the top quartile (top 25\%) of the predicted OBV samples—which better captures the high-impact potential inherent to this role.

\vspace{5pt} 
\noindent \textbf{Case Study 1: Who Performs Better as a Striker — and Does It Change Across Teams?} \\
We investigate how striker performance varies when identical action contexts are assigned to different players by applying our substitution framework to two contrasting tactical systems: Manchester City’s striker role (Haaland’s context) and Manchester United’s striker role (Højlund’s context), both taken from the 2023/24 season. We replace the striker’s embedding with those of several stylistically similar forwards and compare the resulting predicted behavioral tendencies and expected rOBV. The results are summarized in Tables 3 and 4.
\begin{table}[h!]
\centering
\caption{Comparison of simulated striker performance under Haaland’s and Højlund’s tactical contexts. OBV values shown in parentheses indicate each player's average OBV across the 2020/21–2023/24 seasons.}

\label{tab:striker_1_context}

\renewcommand{\arraystretch}{1.2}
\setlength{\tabcolsep}{3pt} 

\resizebox{\linewidth}{!}{
\begin{tabular}{lccccc}
\toprule
\textbf{Player} & 
\makecell{\textbf{Dribble} \\ \textbf{(\%)}} & 
\makecell{\textbf{Pass} \\ \textbf{(\%)}} & 
\makecell{\textbf{Shot} \\ \textbf{(\%)}} & 
\makecell{\textbf{Success} \\ \textbf{Rate (\%)}} & 
\makecell{\textbf{Pred.} \\ \textbf{rOBV $\uparrow$}} \\
\midrule

\makecell[l]{Alexander Isak \\ (OBV: 4.94)}       & \textbf{64.34} & 15.65 & 9.19 & \textbf{81.83} & \textbf{3.76} \\
\makecell[l]{Rasmus Højlund \\ (OBV: 1.80)}       & 61.65 & 17.55 & 9.47 & 79.51 & 2.23 \\
\makecell[l]{Darwin Núñez \\ (OBV: 2.49)}         & 61.82 & 18.01 & 9.64 & 80.57 & 2.12 \\
\makecell[l]{Jean-Philippe Mateta \\ (OBV: 0.65)} & 57.43 & 21.05 & \textbf{10.54} & 78.08 & 1.94 \\

\addlinespace[2pt]
\cdashline{1-6} 
\addlinespace[2pt]

\makecell[l]{Erling Haaland (Sim) \\ (OBV: 4.05)} & 57.30 & \textbf{21.13} & 10.39 & 77.93 & 2.71 \\
\makecell[l]{Erling Haaland (GT)}                 & 59.35 & 18.11 & 12.33 & 78.42 & 2.59 \\
\bottomrule
\end{tabular}
}

\vspace{10pt} 

\resizebox{\linewidth}{!}{
\begin{tabular}{lccccc}
\toprule
\textbf{Player} & 
\makecell{\textbf{Dribble} \\ \textbf{(\%)}} &  
\makecell{\textbf{Pass} \\ \textbf{(\%)}} & 
\makecell{\textbf{Shot} \\ \textbf{(\%)}} & 
\makecell{\textbf{Success} \\ \textbf{Rate (\%)}} & 
\makecell{\textbf{Pred.} \\ \textbf{rOBV $\uparrow$}} \\
\midrule

\makecell[l]{Alexander Isak \\ (OBV: 4.94)}       & \textbf{66.57} & 16.46 & 3.41 & \textbf{85.44} & \textbf{4.65} \\
\makecell[l]{Darwin Núñez \\ (OBV: 2.49)}         & 64.84 & 18.30 & 3.90 & 84.70 & 3.91 \\
\makecell[l]{Erling Haaland \\ (OBV: 4.05)}       & 61.89 & 21.20 & 4.03 & 83.40 & 1.37 \\
\makecell[l]{Jean-Philippe Mateta \\ (OBV: 0.65)} & 60.36 & \textbf{21.30} & \textbf{4.34} & 82.34 & 1.23 \\

\addlinespace[2pt]
\cdashline{1-6} 
\addlinespace[2pt]

\makecell[l]{Rasmus Højlund (Sim) \\ (OBV: 1.80)} & 64.39 & 18.30 & 3.51 & 83.74 & 2.23\\
\makecell[l]{Rasmus Højlund (GT)}                 & 62.22 & 21.56 & 2.26 & 84.39 & 1.67 \\
\bottomrule
\end{tabular}
}
\end{table}
Across Manchester City’s structured, possession-oriented attacking environment (Table 3), the model predicts that Alexander Isak provides the highest contribution when substituted into Haaland’s match contexts. His greater dribble involvement and ability to create separation under pressure are rewarded in situations where City progress play through controlled circulation and final-third overloads. However, when we simulate the same set of forwards in Manchester United’s transition-dependent attacking patterns (Table 4), the ranking changes. Even elite finishers such as Haaland experience a noticeable decline in predicted rOBV when inserted into Højlund’s contexts. Likewise, Højlund’s observed performance appears lower than his underlying ability. This shift reflects structural differences in midfield stability, progression quality, and final-third support, rather than differences in finishing talent.

To examine this effect more systematically, we extend the analysis by cross-substituting each
striker into all other striker contexts as summarized in Table 5.
\begin{table}[h!]
\centering
\caption{Cross-substitution results for elite strikers across different tactical contexts. Each row indicates the substitute player and each column represents the original player’s context. Bold values denote the highest predicted rOBV for each context. This illustrates that optimal player fit varies across tactical environments.}
\label{tab:striker_2_context}

\renewcommand{\arraystretch}{1.25}
\setlength{\tabcolsep}{2pt} 

\resizebox{\linewidth}{!}{
\begin{tabular}{cccccc}
\toprule
\textbf{Player} & 
\makecell{\textbf{Context:} \\ \textbf{Haaland}} & 
\makecell{\textbf{Context:} \\ \textbf{Højlund}} & 
\makecell{\textbf{Context:} \\ \textbf{Isak}} & 
\makecell{\textbf{Context:} \\ \textbf{Núñez}} & 
\makecell{\textbf{Context:} \\ \textbf{Mateta}} \\
\midrule

\makecell[c]{Erling \\ Haaland}       & 2.71 & 1.37 & 5.72 & 3.27 & 2.38 \\
\makecell[c]{Rasmus \\ Højlund}       & 2.23 & 2.23 & 6.19 & 4.17 & 2.48 \\
\makecell[c]{Alexander \\ Isak}       & \textbf{3.76} &  \textbf{4.65} & 9.23 &  \textbf{8.09} &  \textbf{5.42}\\
\makecell[c]{Darwin \\ Núñez}       & 2.12 & 3.91 &  \textbf{10.23} & 7.09 & 3.96 \\
\makecell[c]{Jean-Philippe \\ Mateta} & 1.94 & 1.23 & 4.72 & 2.30 & 1.71 \\

\addlinespace[3pt]
\cdashline{1-6} 
\addlinespace[3pt]

\makecell[c]{Original \\ Player} & 2.59 & 1.67 & 6.98 & -0.33 & 3.24 \\
\bottomrule
\end{tabular}
}
\end{table}
This allows us to observe how the same player’s contribution changes when the surrounding tactical environment shifts. The results reinforce the earlier pattern: Haaland’s effectiveness is highly system-dependent. While he remains productive in Manchester City’s structured buildup, his predicted rOBV declines substantially when placed into the more fragmented and transition-heavy environments associated with Højlund, Núñez, or Mateta. In contrast, Alexander Isak consistently maintains strong performance across multiple contexts. His ability to generate value without relying on high-frequency service makes him more adaptable to varied attacking structures. Interestingly, Núñez and Isak appear to benefit from being placed into each other’s contexts.

We also note that Núñez’s ground-truth rOBV in the 2023/24 season was unusually low. The discrepancy between his observed performance (GT) and the model’s predicted outcomes in certain contexts should therefore be interpreted as a limitation of the input season rather than a miscalibration of player ability. In other words, the model is sensitive to the performance distribution present in the training window, and drops in observed OBV can propagate into the baseline against which substitutions are compared.

\vspace{5pt} 
\noindent \textbf{Case Study 2: Who Generates the Most Value in Bukayo Saka's Role?}\\
Whereas the preceding experiment demonstrated system dependency, this case study applies the model to a large-scale scouting scenario. We evaluated all right-wingers with over 1500 minutes played (from the 2020/21-2024/25 seasons) within the tactical context of Bukayo Saka's 2023/24 season to identify the optimal performers.

First, to establish a fidelity benchmark, Saka was simulated within his own context. As shown in Table 6, the predicted rOBV of Saka (18.59) was notably higher than his actual value (15.72). This positive bias suggests the model identifies a higher potential value ceiling for this role than was captured in the ground truth. Therefore, the simulated rOBV (18.59) is used as the baseline for all subsequent comparisons.
\begin{table}[h!]
\Large
\centering
\caption{Top predicted right-winger alternatives under Bukayo Saka’s 2023/24 tactical contexts. Among all right wingers in the league, the players shown are those with the highest predicted rOBV when substituted into Saka’s match contexts.}
\label{tab:striker_3_context}

\renewcommand{\arraystretch}{1.2}
\setlength{\tabcolsep}{3pt} 

\resizebox{\linewidth}{!}{
\begin{tabular}{ccccccc}
\toprule
\textbf{Player} & 
\makecell{\textbf{Dribble} \\ \textbf{(\%)}} & 
\makecell{\textbf{Pass} \\ \textbf{(\%)}} & 
\makecell{\textbf{Shot} \\ \textbf{(\%)}} & 
\makecell{\textbf{Defensive} \\ \textbf{Actions(\%)}} & 
\makecell{\textbf{Success} \\ \textbf{Rate(\%)}} &
\makecell{\textbf{Pred.} \\ \textbf{rOBV} $\uparrow$} \\
\midrule

\makecell[c]{Mohamed Salah \\ (OBV: 10.96)}       & 68.67 & 10.68 & 1.01 & 6.33 & 90.18 & 19.78 \\
\makecell[c]{Noni Madueke \\ (OBV: 3.70)}         & 70.91 & 6.78 & 1.08 & 6.44 & 91.49 & 19.36 \\
\makecell[c]{Antoine Semenyo \\ (OBV: 2.46)}      & 71.25 & 6.25 & 1.10 & 6.34 & 90.29 & 19.22 \\
\makecell[c]{Adama Traor\'{e} \\ (OBV: 10.96)}       & 73.47 & 5.44 & 0.73 & 5.85 & 91.29 & 18.92 \\

\addlinespace[3pt]
\cdashline{1-7} 
\addlinespace[3pt]

\makecell[c]{Bukayo Saka (Sim) \\  (OBV: 11.31)} & 71.19 & 6.31 & 0.97 & 6.31 & 86.16 & 18.59 \\
\makecell[c]{Bukayo Saka (GT)} & 66.90 & 9.86 & 0.62 & 6.95 & 84.24 & 15.72 \\
\bottomrule
\end{tabular}
}
\end{table}

Among all evaluated players, Mohamed Salah achieves the highest predicted rOBV (19.78). In Saka’s context, the model projects Salah to adopt a more pass-inclusive chance creation profile, reflected in his elevated Pass (\%) and high Success Rate (\%). This suggests that Salah could maintain value not only through dribble-led progression but also through controlled, playmaking-oriented choices—representing a different but equally valid interpretation of the role.

Noni Madueke follows closely (19.36), with a profile centered on high dribble involvement and efficient decisive touches, similar to Saka’s own style. Antoine Semenyo and Adama Traoré also generate high predicted rOBV under these conditions, though with stylistic variation: Semenyo projects a slightly more balanced progression approach, while Traoré represents the extreme isolation-driven dribbler archetype, relying heavily on individual ball-carrying to advance play. 

Taken together, these results indicate that Saka’s tactical environment amplifies players who can consistently win isolated duels and advance possession without heavy passing support. Notably, while these players differ in how they balance dribbling vs. passing, the shared requirement across high-performers is self-generated progression with a high success rate.

\vspace{5pt} 
\noindent \textbf{Case Study 3: Can the Model Identify Suitable Alternatives for Bukayo Saka’s Role?}\\
In this case study, we examine whether the model can autonomously identify viable stylistic alternatives for Bukayo Saka, without relying on manually selected candidates. Players were retrieved directly from the learned player embedding space using cosine similarity, and each candidate was substituted into Saka’s match contexts to evaluate how their behavior and expected value would translate under identical tactical conditions. The results are presented in Table 7.

The model predicts that both Mohamed Salah and Noni Madueke would deliver the strongest performance under Saka’s tactical environment. Both players reproduce Saka’s core high-dribble wing profile, but with more efficient execution: Salah shows a notably higher pass frequency (10.68\%), while Madueke maintains an even higher dribble share (70.91\%) paired with the highest estimated success rate in the cohort (91.49\%). These findings suggest that both players could assume Saka’s progression-focused right-wing role while preserving its creative threat profile. Riyad Mahrez also emerges as a plausible stylistic substitute, though with a slightly lower predicted rOBV (17.50). His projected action distribution remains close to Saka’s simulated baseline, but with more dribble-dominant possession sequences (71.26\% dribble rate), reflecting a more self-creation-oriented approach to chance generation.

In contrast, Bryan Mbeumo and Jarrod Bowen represent instructive negative cases. Although they display competitive action efficiency, both are predicted to generate substantially lower rOBV (11.80 and 10.10, respectively). Notably, Mbeumo’s shot selection rate (1.26\%) indicates a more direct finishing-leaning approach, while Bowen’s moderate pass/dribble split suggests reduced ability to sustain ball progression in isolated wide-channel situations. In Saka’s context, where high-pressure self-progression is structurally required, the model projects that these players would contribute less value.
\begin{table}[h!]
\Large
\centering
\caption{Comparison of simulated right-winger performance under Bukayo Saka’s tactical context. Players listed with parentheses denote their embedding-similarity rank relative to Saka, based on the learned player embedding space.}
\label{tab:striker_4_context}

\renewcommand{\arraystretch}{1.2}
\setlength{\tabcolsep}{3pt} 

\resizebox{\linewidth}{!}{
\begin{tabular}{ccccccc}
\toprule
\textbf{Player} & 
\makecell{\textbf{Dribble} \\ \textbf{(\%)}} & 
\makecell{\textbf{Pass} \\ \textbf{(\%)}} & 
\makecell{\textbf{Shot} \\ \textbf{(\%)}} & 
\makecell{\textbf{Defensive} \\ \textbf{Actions(\%)}} & 
\makecell{\textbf{Success} \\ \textbf{Rate(\%)}} &
\makecell{\textbf{Pred.} \\ \textbf{rOBV} $\uparrow$} \\
\midrule

\makecell[c]{Mohamed Salah(4)}       & 68.67 & 10.68 & 1.01 & 6.33 & 90.18 & 19.78 \\
\makecell[c]{Noni Madueke(5)}         & 70.91 & 6.78 & 1.08 & 6.44 & 91.49 & 19.36 \\
\makecell[c]{Riyad Mahrez(16)}      & 71.26 & 6.50 & 1.00 & 6.27 & 87.56 & 17.50 \\
\makecell[c]{Bryan Mbeumo(7)}       & 67.59 & 9.32 & 1.26 & 6.78 & 83.76 & 11.80 \\

\addlinespace[3pt]
\cdashline{1-7} 
\addlinespace[3pt]

\makecell[c]{Bukayo Saka (Sim) \\  (OBV: 11.31)} & 71.19 & 6.31 & 0.97 & 6.31 & 86.16 & 18.59 \\
\makecell[c]{Bukayo Saka (GT)} & 66.90 & 9.86 & 0.62 & 6.95 & 84.24 & 15.72 \\
\bottomrule
\end{tabular}
}
\end{table}

Taken together, these results demonstrate that embedding-based retrieval combined with counterfactual substitution provides a quantitative method for evaluating role suitability. Players who appear superficially similar when viewed statistically or stylistically may diverge sharply in their expected contribution once evaluated within the tactical and positional demands of a specific match context.

\vspace{5pt} 
\noindent \textbf{Case Study 4: Does High On-Ball Value Transfer Across Positions?}\\
To further illustrate that high on-ball value does not necessarily generalize across roles, we conduct an additional counterfactual experiment in which Erling Haaland, a high-impact striker, is substituted into four defensive roles within Arsenal’s tactical context. While Haaland’s ground-truth rOBV as a striker is among the highest in the league, the model predicts a substantial decline in his contribution when he is placed into defensive match contexts. In each substitution scenario, the original Arsenal defenders outperform Haaland, achieving both higher predicted rOBV and more role-appropriate action profiles. The full results of this experiment are summarized in Table 8.

For example, when substituted for Oleksandr Zinchenko or Gabriel Magalhães, Haaland’s predicted action distribution shifts markedly toward low-impact passing with reduced on-ball involvement, resulting in lower projected value compared to the original defenders. A similar pattern emerges when replacing William Saliba or Ben White: although Haaland retains his typical offensive efficiency, his limited defensive-action frequency and reduced contribution to buildup play lead to an overall decrease in expected value within these roles.
\begin{table}[h!]
\Large
\centering
\caption{Counterfactual substitution of Erling Haaland into Arsenal defensive contexts. The first row shows the model-predicted rOBV when Haaland replaces the original defender in each context. “Original Player (Sim)” denotes the model’s simulation of the defender in their own context, and “Original Player (GT)” reports the observed rOBV from match data.}
\label{tab:striker_5_context}

\renewcommand{\arraystretch}{1.2}
\setlength{\tabcolsep}{3pt} 

\resizebox{\linewidth}{!}{%
    \begin{tabular}{ccccc}
    \toprule
    \textbf{Player} & 
    \makecell{\textbf{Context:} \\ \textbf{Oleksandr} \\
             \textbf{Zinchenko} \\ (OBV: 4.34)}  & 
    \makecell{\textbf{Context:} \\ \textbf{Gabriel} \\
             \textbf{Magalhães} \\ (OBV: 2.98)}  & 
    \makecell{\textbf{Context:} \\ \textbf{William} \\
             \textbf{Saliba} \\ (OBV: 3.68)}  & 
    \makecell{\textbf{Context:} \\ \textbf{Ben} \\
             \textbf{White} \\ (OBV: 6.13)}  \\
    
    \midrule
    
    \makecell[c]{Erling  \\ Haaland}       & 2.35 & 1.42 & 1.98 & 1.37 \\
    
    \addlinespace[3pt]
    \cdashline{1-5} 
    \addlinespace[3pt]
    
    \makecell[c]{Original \\ Player(Sim)} & 5.19 & 3.63 & 5.14 & 8.78 \\
    \makecell[c]{Original \\ Player(GT)} & 4.79 & 4.00 & 5.23 & 7.74 \\
    \bottomrule
    \end{tabular}%
}
\end{table}
Importantly, the model is not given positional labels, so it does not assume that Haaland “should” play as a striker or that the original players are defenders. Instead, Haaland is simply substituted into the same match contexts the original players faced. Because of this, the model does not penalize him for being in a different nominal position. However, his predicted rOBV still changes depending on the contextual demands of the role. This shows that performance is shaped by context, not by positional labels, and that role fit matters when evaluating player suitability.

\section{Conclusions}

We present a player-conditioned and value-aware approach to modeling football match events as autoregressive sequences, enabling the estimation of how a player’s on-ball contribution depends on surrounding tactical context. By jointly predicting event attributes and residual on-ball value, the model captures both the immediate structure of play and the longer-term impact of individual decisions. The approach outperforms sequence-based baselines in next-event prediction accuracy, spatial precision, and future contribution estimation, and the learned player embeddings exhibit interpretable role structure without relying on positional supervision. Furthermore, hypothetical player substitution allows the evaluation of transfer fit by quantifying how a player’s projected contribution would change in a new tactical environment. Future work will incorporate tracking data and player spatial trajectories to better represent off-ball behavior and positional interactions, and will extend the generative framework to full-possession or match-level simulation to support broader tactical and recruitment decision-making.


\bibliographystyle{ACM-Reference-Format}
\bibliography{HUDL}

@misc{FIFA2025Transfers,
  author       = {{FIFA}},
  title        = {Global Transfer Market Report: Mid-Year 2025},
  year         = {2025},
  howpublished = {\url{https://inside.fifa.com/transfer-system/media-releases/global-transfer-market-new-all-time-highs-2025-mid-year-window}},
  note         = {Accessed: 2025-07-19}
}

@misc{Singh2019xT,
  author       = {Singh, Karun},
  title        = {Expected Threat},
  year         = {2019},
  howpublished = {\url{https://karun.in/blog/expected-threat.html}},
  note         = {Accessed: 2025-07-19}
}

@misc{StatsBomb2021OBV,
  author       = {{StatsBomb}},
  title        = {Introducing On-Ball Value (OBV)},
  year         = {2021},
  howpublished = {\url{https://blogarchive.statsbomb.com/news/introducing-on-ball-value-obv/}},
  note         = {Accessed: 2025-07-19}
}

@misc{Karpathy2022nanoGPT,
  author       = {Karpathy, Andrej},
  title        = {nanoGPT},
  year         = {2022},
  publisher    = {GitHub},
  journal      = {GitHub repository},
  howpublished = {\url{https://github.com/karpathy/nanoGPT}},
  note         = {Accessed: 2025-07-19}
}

@article{Berlinschi2013,
  author    = {Berlinschi, Ruxanda and Schokkaert, Jeroen and Swinnen, Johan},
  title     = {When Drains and Gains Coincide: Migration and International Football Performance},
  journal   = {Labour Economics},
  volume    = {21},
  pages     = {1--14},
  year      = {2013},
  doi       = {10.1016/j.labeco.2012.12.006}
}

@article{Jarjabka2024,
  author    = {Jarjabka, {\'A}d{\'a}m and F{\H{u}}r{\'e}sz, D{\'a}vid Istv{\'a}n and Havran, Zolt{\'a}n},
  title     = {The Impact of Cultural Distance on the Migration of Professional Athletes as High-Skilled Employees},
  journal   = {Journal of Industrial and Business Economics},
  volume    = {51},
  pages     = {585--603},
  year      = {2024},
  doi       = {10.1007/s40812-023-00288-8}
}

@inproceedings{DecroosVAEP2021,
  author    = {Decroos, Tom and Bransen, Lotte and Van Haaren, Jan and Davis, Jesse},
  title     = {VAEP: An Objective Approach to Valuing On-the-Ball Actions in Soccer (Extended Abstract)},
  booktitle = {Proceedings of the 29th International Joint Conference on Artificial Intelligence (IJCAI)},
  year      = {2020},
  pages     = {4696--4700},
  doi       = {10.24963/ijcai.2020/648}
}

@inproceedings{Fernandez2019EPV,
  author    = {Fern{\'a}ndez, Javier and Bornn, Luke and Cervone, Dan},
  title     = {Decomposing the Immeasurable Sport: A Deep Learning Expected Possession Value Framework for Soccer},
  booktitle = {Proceedings of the 13th MIT Sloan Sports Analytics Conference},
  year      = {2019}
}

@article{Pappalardo2019,
  author    = {Pappalardo, Luca and Cintia, Paolo and Ferragina, Paolo and Massucco, Emanuele and Pedreschi, Dino and Giannotti, Fosca},
  title     = {PlayeRank: Data-driven Performance Evaluation and Player Ranking in Soccer via a Machine Learning Approach},
  journal   = {ACM Transactions on Intelligent Systems and Technology},
  volume    = {10},
  number    = {5},
  year      = {2019},
  pages     = {1--27},
  doi       = {10.1145/3343172}
}

@inproceedings{DecroosVector2019,
  author    = {Decroos, Tom and Davis, Jesse},
  title     = {Player Vectors: Characterizing Soccer Players’ Playing Style from Match Event Streams},
  booktitle = {Machine Learning and Knowledge Discovery in Databases (ECML PKDD)},
  series    = {Lecture Notes in Computer Science},
  volume    = {11908},
  year      = {2020},
  pages     = {569--584},
  doi       = {10.1007/978-3-030-46133-1_34}
}

@inproceedings{YilmazEmbedding2022,
  author    = {Y{\i}lmaz, {\"O}znur {\.I}layda and {\"O}{\u{g}}{\"u}d{\"u}c{\"u}, {\c{S}}ule G{\"u}nd{\"u}z},
  title     = {Learning Football Player Features Using Graph Embeddings for Player Recommendation System},
  booktitle = {Proceedings of the 37th ACM/SIGAPP Symposium on Applied Computing},
  year      = {2022},
  pages     = {577--584},
  doi       = {10.1145/3477314.3507257}
}

@inproceedings{Rovshitz2024TransformerSoccer,
  author    = {Rovshitz, Aviv and Puzis, Rami},
  title     = {Transformer-Based Framework for Versatile Analysis of Events Data in Soccer},
  booktitle = {Proceedings of the ECML/PKDD 2024 Workshop on Machine Learning and Data Mining for Sports Analytics (MLSA 2024)},
  year      = {2024}
}

@misc{adjileye2024risingballer,
  author    = {Adjileye, Akedjou Achraff},
  title     = {RisingBALLER: A Player is a Token, a Match is a Sentence, A Path Towards a Foundational Model for Football Players Data Analytics},
  year      = {2024},
  eprint    = {2410.00943},
  archivePrefix = {arXiv},
  primaryClass = {cs.LG},
  url       = {https://arxiv.org/abs/2410.00943}
}

@inproceedings{Simpson2022Seq2event,
  author    = {Simpson, Ian and Beal, Ryan J. and Locke, Duncan and Norman, Timothy J.},
  title     = {Seq2Event: Learning the Language of Soccer Using Transformer-based Match Event Prediction},
  booktitle = {Proceedings of the 28th ACM SIGKDD Conference on Knowledge Discovery and Data Mining},
  year      = {2022},
  pages     = {3898--3908},
  doi       = {10.1145/3534678.3539138}
}

@article{Yeung2025NMSTPP,
  author    = {Yeung, Calvin and Sit, Tony and Fujii, Keisuke},
  title     = {Transformer-based Neural Marked Spatio-Temporal Point Process Model for Analyzing Football Match Events},
  journal   = {Applied Intelligence},
  volume    = {55},
  number    = {5},
  year      = {2025},
  doi       = {10.1007/s10489-024-05996-9}
}

@inproceedings{MendesNeves2026LEM,
  author    = {Mendes-Neves, Tiago and Meireles, Lu{\'i}s and Mendes-Moreira, Jo{\~a}o},
  title     = {A Scalable Approach for Unified Large Events Models in Soccer},
  booktitle = {Machine Learning and Knowledge Discovery in Databases (ECML PKDD)},
  series    = {Lecture Notes in Computer Science},
  volume    = {16022},
  year      = {2026},
  doi       = {10.1007/978-3-032-06129-4_21}
}

\appendix



\end{document}